\documentclass{article} 
\usepackage{iclr2021_conference,times}
\usepackage{booktabs}

\input{math_commands.tex}

\usepackage{hyperref}
\usepackage{url}
\usepackage[linesnumbered,ruled,vlined]{algorithm2e}

\usepackage{amsthm}
\usepackage{latexsym}
\usepackage{graphicx}
\usepackage{subfigure} 
\usepackage{url}
\usepackage{enumerate}
\usepackage{amsmath}
\usepackage{mathtools}
\usepackage{amssymb}

\usepackage{tabularx, booktabs, enumitem}
\newcolumntype{Y}{>{\centering\arraybackslash}X}
\newcolumntype{L}{>{\arraybackslash}X}
\newcolumntype{R}{>{\raggedleft\arraybackslash}X}

\usepackage{multirow}
\usepackage{xspace}

\usepackage{booktabs} 

\def \our {\mbox{APART}\xspace}
\def \fgsm {\mbox{\tiny FGSM}}
\def \fgsmp {\mbox{\tiny FGSM+}}
\def \random {\mbox{\tiny Random}}
\newcommand{\pgd}[1]{\mbox{\tiny PGD-#1}}
\newcommand{\cnn}{\mathop{\mathrm{CNNs}}}
\newcommand{\smallsection}[1]{\noindent\textbf{#1.}}
\newcommand{\tablespace}{\vspace{0.1cm}}

\newcommand{\ie}{\textit{i.e.}}
\newcommand{\eg}{\textit{e.g.}}
\newcommand{\st}{\textit{s.t.}}

\title{Overfitting or Underfitting? Understand\\Robustness Drop in Adversarial Training}

\iclrfinalcopy
\author{
Zichao Li\thanks{Equal Contributions.}\;\,\textsuperscript{$\dagger$}~~
Liyuan Liu$^{*}$\textsuperscript{$\mathsection$}~~
Chengyu Dong\textsuperscript{$\dagger$}~~
Jingbo Shang\textsuperscript{$\dagger$}\\
\textsuperscript{$\dagger$}\texttt{\small \{zil023, cdong, jshang\}@ucsd.edu}\;\;\;\textsuperscript{$\mathsection$}\texttt{\small ll2@illinois.edu}
\\
\textsuperscript{$\dagger$}{University of California San Diego} \;\;\;\textsuperscript{$\mathsection$}{University of Illinois at Urbana-Champaign}
}

\begin{document}
\maketitle

\begin{abstract}
Our goal is to understand \emph{why the robustness drops after conducting adversarial training for too long}.
Although this phenomenon is commonly explained as overfitting, our analysis suggest that its primary cause is \emph{perturbation underfitting}. 
We observe that after training for too long, FGSM-generated perturbations deteriorate into random noise.
Intuitively, since no parameter updates are made to strengthen the perturbation generator, once this process collapses, it could be trapped in such local optima.
Also, sophisticating this process could mostly avoid the robustness drop, which supports that this phenomenon is caused by underfitting instead of overfitting. 
In the light of our analyses, we propose \our, an \underline{a}da\underline{p}tive \underline{a}dve\underline{r}sarial \underline{t}raining framework, which parameterizes perturbation generation and progressively strengthens them.
Shielding perturbations from underfitting unleashes the potential of our framework. 
 In our experiments, \our provides comparable or even better robustness than PGD-10, with only about $1/4$ of its computational cost.
\end{abstract}

\section{Introduction}

While neural networks keep advancing the state of the arts,
their vulnerability to adversarial attacks casts a shadow over their applications---subtle, human-imperceptible input shifts can fool these models and alter their predictions. 
To establish model robustness, adversarial training adds perturbations to model inputs during training and is considered one of the most successful approaches.
However, it leads to an unexpected phenomenon---after conducting adversarial training for too long, the model robustness starts to drop, and in some cases, even diminishes entirely. 
Our objective is to explore and address the underlying issue of this robustness drop phenomenon.

The robustness drop has been widely viewed as overfitting.
However, sophisticated methods like PGD-10 have a smaller drop than simple methods like FGSM\footnote{On the CIFAR-10 dataset, for a FGSM trained Pre-ResNet18 model, its accuracy under PGD-20 attack drops from 44.36\% (the 20th epoch) to 0\% (the 21st epoch); For a PGD-10 trained model, it drops from 47.43\% (the 30th epoch) to 31.71\% (the 200th epoch). }, making this phenomenon more similar to underfitting than overfitting.
Here, we systematically analyze the robustness drop and recognize its primary cause as \emph{perturbation underfitting}, an important but long-overlooked issue. 

\begin{figure*}[t]
 
    \subfigure[FGSM-Generated Perturbations for Pre-ResNet18.]{
        \label{fig:random}
        \includegraphics[width=0.52\textwidth]{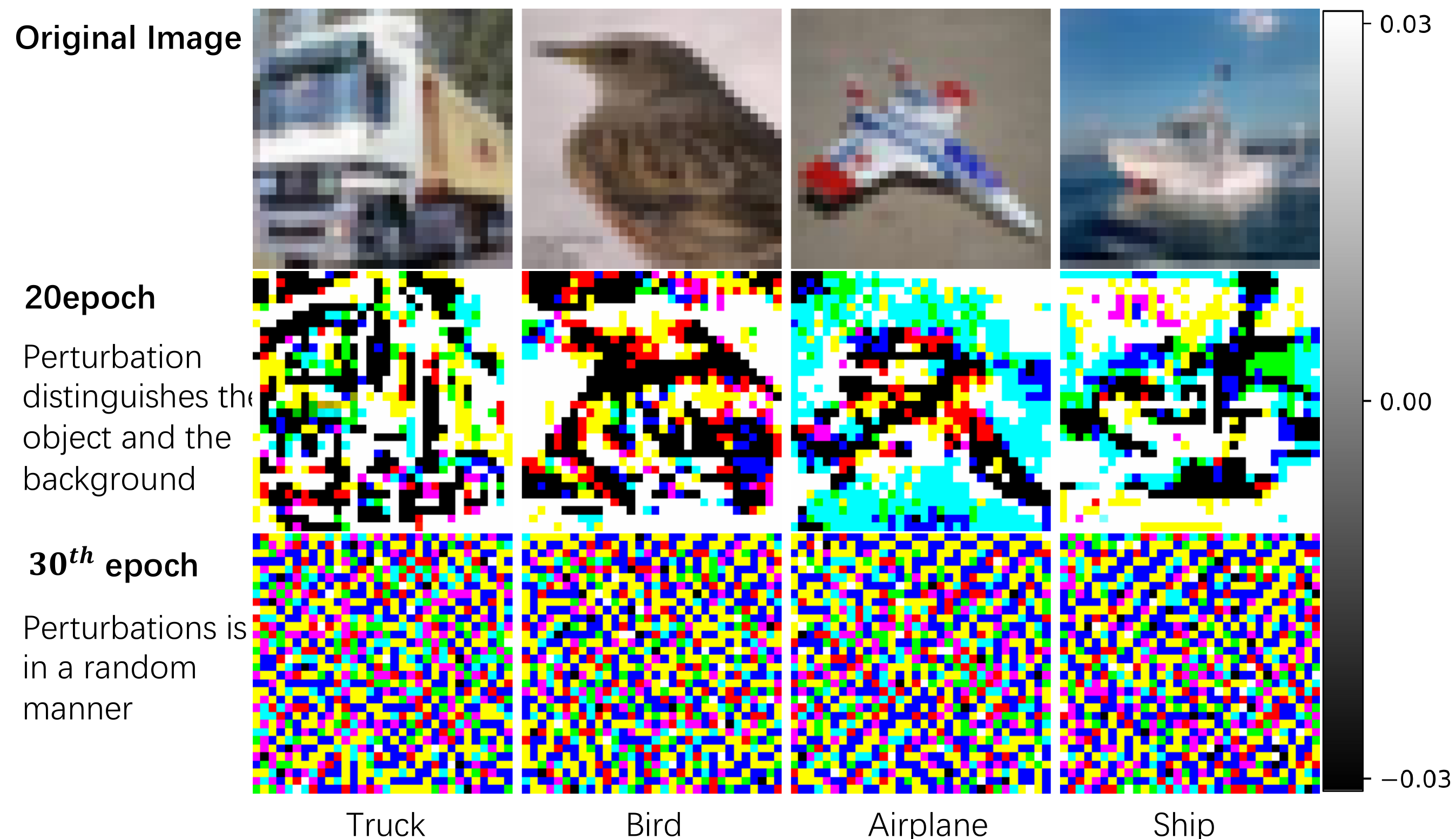}
    } 
	\subfigure[Test Accuracy of Pre-ResNet18 on CIFAR-10.
	]{
	    \label{fig:defense}
	    \includegraphics[width=0.46\textwidth]{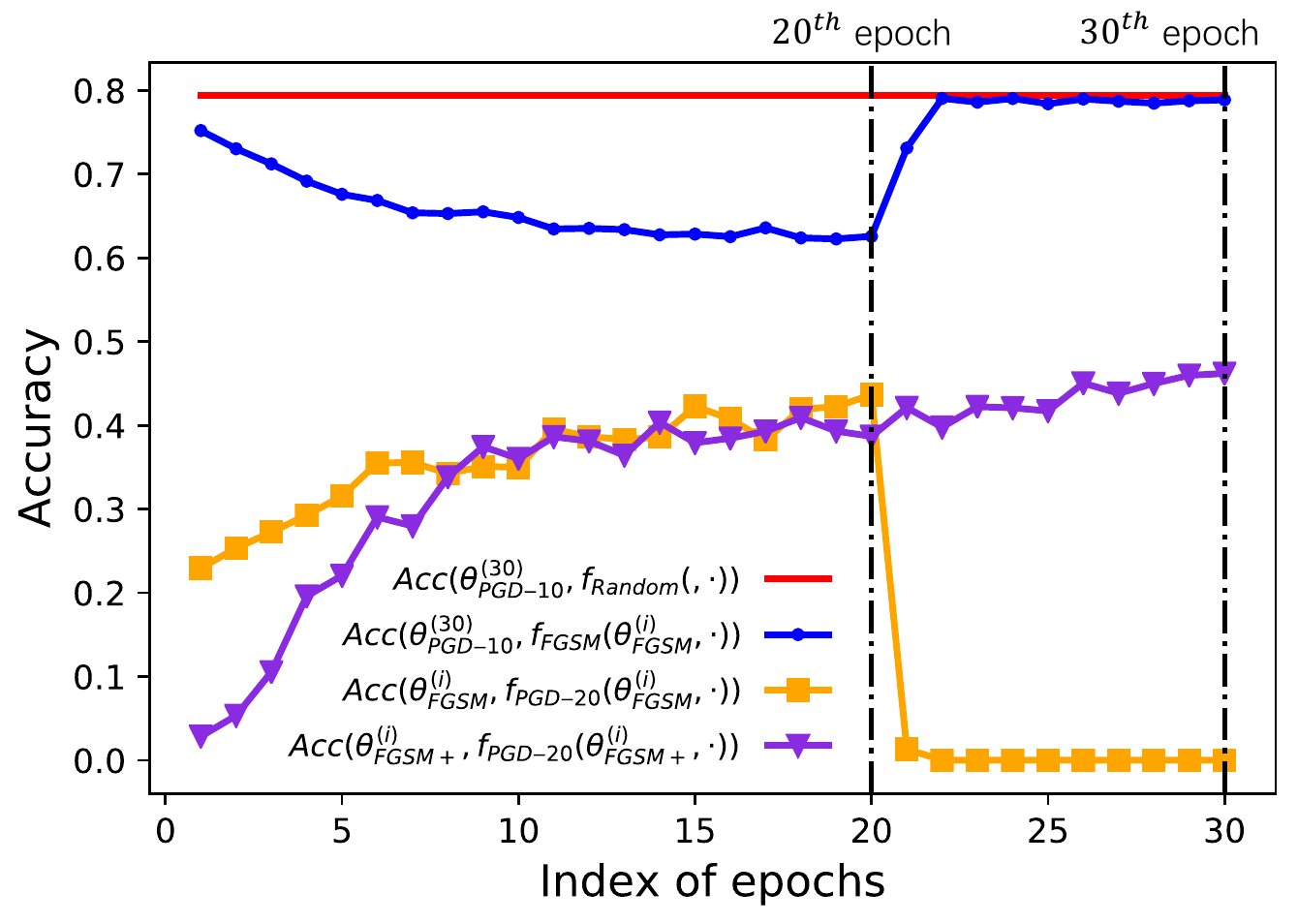}
	}
   \caption{
   Analysis of FGSM-generated perturbations. The notation $Acc(\cdot)$ is elaborated in Table~\ref{table:notation}.
   }
   \label{fig:fgsm_analysis}
\end{figure*}

Firstly, along with the robustness drop, we observe a consistent decrease of perturbation strength. 
As visualized in Figure~\ref{fig:random}, although FGSM-generated perturbations can surgically doctor the image at the 20th epoch, they deteriorate into random noise at the 30th epoch.
We also estimate perturbation strength at different epochs with a separately trained model. 
As in Figure~\ref{fig:defense}, the strength of FGSM-generated perturbations suddenly drops to the random noise level at the 21st epoch.
Note that the 21st epoch is the same epoch when the robustness drop happens, which is another evidence verifying that the lack of perturbation strength causes the robustness drop. 

We further examine the mechanism behind the robustness drop.
Adversarial training  optimizes $\cL(\cdot)$ as below, and different methods generate perturbations differently ($\theta$ is the network parameter, $\delta$ is the perturbation, $(\xb,y)$ is a data-label pair, and $f(\cdot)$ is the perturbation generation function).
\begin{equation}
\label{eq:min}
    \min_{\theta} \cL(\theta, \xb + \delta, y)\;\st\;\delta=f(\theta, \xb, y).
\end{equation}
Ideally, adversarial training should use the worst-case (i.e., strongest) perturbations, i.e., $f^*(\cdot)=\argmax_{||\delta|| \leq \epsilon}\cL(\theta, \xb+\delta,y)$. 
In practice, as an approximation, $f(\cdot)$ is typically implemented as gradient ascent with fixed iteration and step size.
We define the gap between $f^*(\cdot)$ and $f(\cdot)$ as
\begin{align}
\label{eq:gap}
    \cG(\theta, f, \xb, y) =\max_{||\delta|| \leq \epsilon}\cL(\theta, \xb+\delta,y) - \gL(\theta, \xb+f(\theta, \xb, y),y).
\end{align}
Intuitively, adversarial training cannot establish satisfying model robustness without strong enough perturbations, and strong perturbations require a small gap.
However, to minimize $\gL(\cdot)$, model updates have a tendency to enlarge the gap---weakening the generator would benefit the optimization of Equation~\ref{eq:min} at the cost of larger $\cG(\cdot)$.
Also, since all parameter updates are made to decrease $\gL$, most existing methods have no regularization to close the gap $\cG(\cdot)$.
Thus, FGSM is trapped in the local optima that its perturbations deteriorate into random noise.

From this perspective, the key to prevent the robustness drop falls upon shielding models from perturbation underfitting.
Correspondingly, instead of only updating model parameters and minimizing the objective, we parameterize perturbation generation and update its parameters to maximize the objective and strengthen the perturbation generator. 
First, we treat perturbations as parameters---instead of starting from scratch every time, we update the initialization with gradient ascent.
Also, we factorize the input perturbation as a series of perturbations and then employ learnable step sizes to achieve effective combinations and persistent improvements.
These designs shield our proposed method, \our, from the robustness drop and unleash its potential to perform better.
While \our is about 4 times faster than PGD-10, it achieves comparable or better performance than the state of the art on the CIFAR-10, CIFAR-100, and ImageNet datasets.

\begin{table}[t]
\caption{Notation Table (Elaborated in Section~\ref{sec:prelim})}
\tablespace
\centering
\label{table:notation}
\small
\begin{tabular}{rlrlrlrlrlrl}
\toprule
\multicolumn{2}{c}{$\xb$ is input} & \multicolumn{2}{c}{$y$ is label} & \multicolumn{2}{c}{$\alpha$ is step size} & \multicolumn{2}{c}{$\cL$ is objective} & \multicolumn{2}{c}{$\Delta \xb = \partial \cL / \partial \xb$} & \multicolumn{2}{c}{$\Delta \theta = \partial \cL / \partial \theta$} \\
\midrule
\multicolumn{3}{r}{$f_{\cA}(\theta, \xb, y)$} & \multicolumn{9}{l}{perturbation generation method $\cA$ for $\theta$, $\xb$, $y$} \\ \midrule
\multicolumn{3}{r}{$\theta^{(i)}_{\cA}$} & \multicolumn{9}{l}{model parameter $\theta$ trained for $i$ epochs by method $\cA$ } \\ \midrule
\multicolumn{3}{r}{$\omega_\xb$} & \multicolumn{9}{l}{start point used by the perturbation generation for $\xb$} \\ \midrule
\multicolumn{3}{r}{$\delta^{(i)}_{\cA}$} & \multicolumn{9}{l}{$f_{\cA}(\theta_\cA^{(i)},\cdot)$, \ie, perturbations from method $\cA$ for its $i$-th model $\theta_\cA^{(i)}$} \\ \midrule
\multicolumn{3}{r}{$\cG_{\cA \to \cB} (\theta_\cA^{(i)})$} &  \multicolumn{9}{l}{strength gap between $f_\cA(\theta_\cA^{(i)})$ and $f_\cB(\theta_\cA^{(i)})$ on model $\theta_\cA^{(i)}$}\\ \midrule
\multicolumn{3}{r}{$Acc(\theta^{(i1)}_{\cA}, f_{\cB}(\theta^{(i2)}_{\cC}, \cdot))$} & \multicolumn{9}{l}{accuracy of $\theta^{(i1)}_{\cA}$ under perturbations generated by $f_{\cB}(\cdot)$, for $\theta^{(i2)}_{\cC}$}\\
\bottomrule
\end{tabular}
\end{table}

\section{Preliminaries and Notations}
\label{sec:prelim}

Given a neural network with $n$ convolution blocks, we denote the input of the $i$-th block as $\xb_i$, the input and output of the entire network as $\xb$ and $y$.
Note that, $\xb$ and $\xb_1$ are the same in conventional residual networks, such as ResNet~\citep{he2016deep}, Wide ResNet~\citep{Zagoruyko2016WideRN}, and Pre-Act ResNet~\citep{he2016identity}.
For simplicity, we use $\Delta \xb_i$ and $\Delta \theta$ to denote $\frac{\partial \cL}{\partial \xb_i}$ and $\frac{\partial \cL}{\partial \theta}$, respectively, where $\cL(\theta, \xb, y)$ is the loss function.
As previously mentioned, different adversarial learning algorithms adopt different perturbation generation functions  $f(\cdot)$ but minimize the same objective function as in Equation~\ref{eq:min}.
For example, the FGSM algorithm directly calculates perturbations as $\delta_{\fgsm} = f_{\fgsm}(\theta_{\fgsm}, \xb, y) = \epsilon \cdot \sign(\Delta \xb)$;
the PGD-N algorithm calculates perturbations recursively ($\Pi$ is the projection operator, $\cS$ is the feasible set for perturbations and $f_{\pgd{0}}(\cdot) = 0$):
$$
\delta_{\pgd{N}} = f_{\pgd{N}}(\theta_{\pgd{N}}, \cdot) =  \Pi_{\xb + \cS} \Big(\xb + f_{\pgd{(N-1)}}(\theta_{\pgd{N}}, \cdot) + \epsilon \Delta \big(\xb + f_{\pgd{(N-1)}}(\theta_{\pgd{N}}, \cdot) \big) \Big) - \xb.
$$
For simplicity, we will mostly use $\delta_{\mbox{\scriptsize method}}$ to differentiate different adversarial training algorithms. 
Moreover, we use $\theta^{(i)}_{\mbox{\scriptsize method}}$ to refer to model parameters that are trained by a specific adversarial training algorithm for $i$ epochs, and $\delta^{(i)}_{\mbox{\scriptsize method}}$ to mark the perturbation corresponding to $\theta^{(i)}_{\mbox{\scriptsize method}}$. 
We use $Acc(\theta^{(i1)}_{\cA}, f_{\cB}(\theta^{(i2)}_{\cC}, \cdot))$ to indicate the performance of model $\theta^{(i1)}_{\cA}$, under $f_{\cB}(\theta^{(i2)}_{\cC}, \cdot))$, \ie, the perturbation generated by method $f_{\cB}(\cdot)$ to attack $\theta^{(i2)}_{\cC}$.
These notations are summarized in Table~\ref{table:notation}. 
\section{From the Robustness Drop to the Perturbation Strength}
\label{sec:overfit}

\begin{figure*}[t]
    \subfigure[Training Accuracy]{\includegraphics[width=0.45\textwidth]{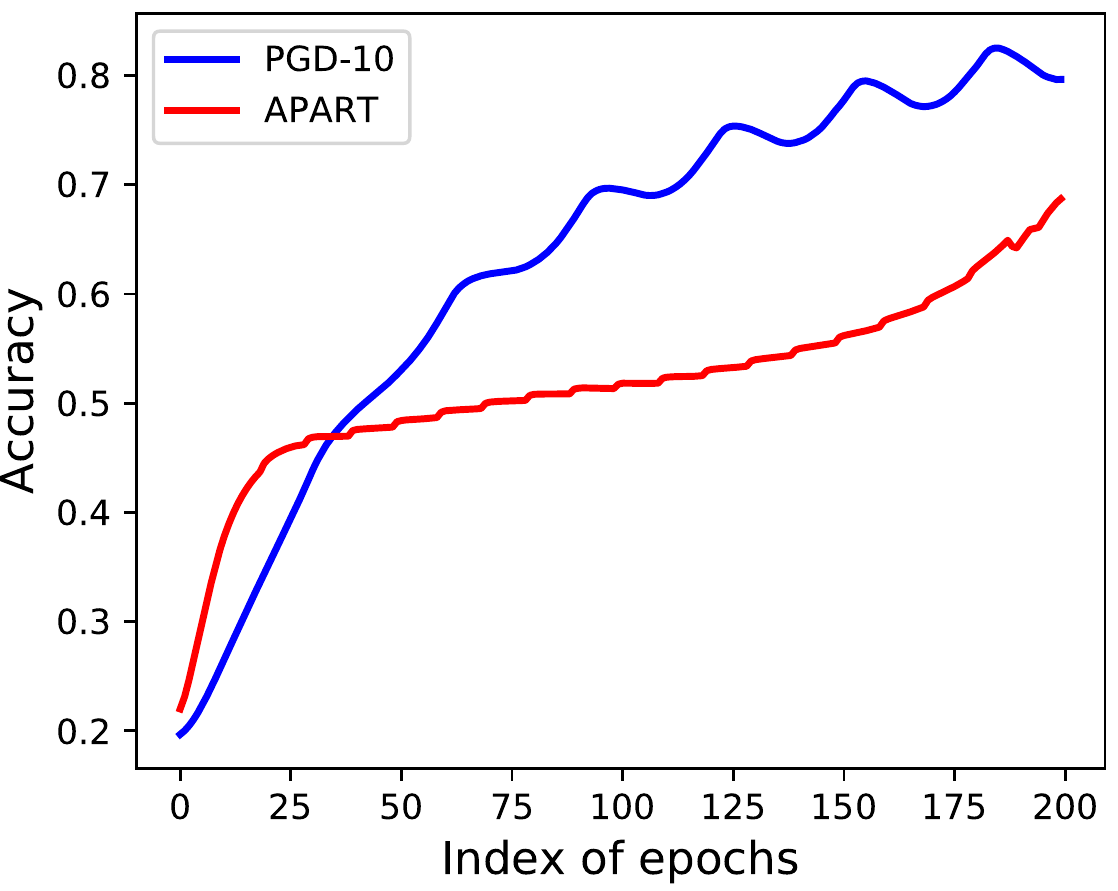}} 
	\subfigure[Test Accuracy under PGD-20 Attack]{\includegraphics[width=0.45\textwidth]{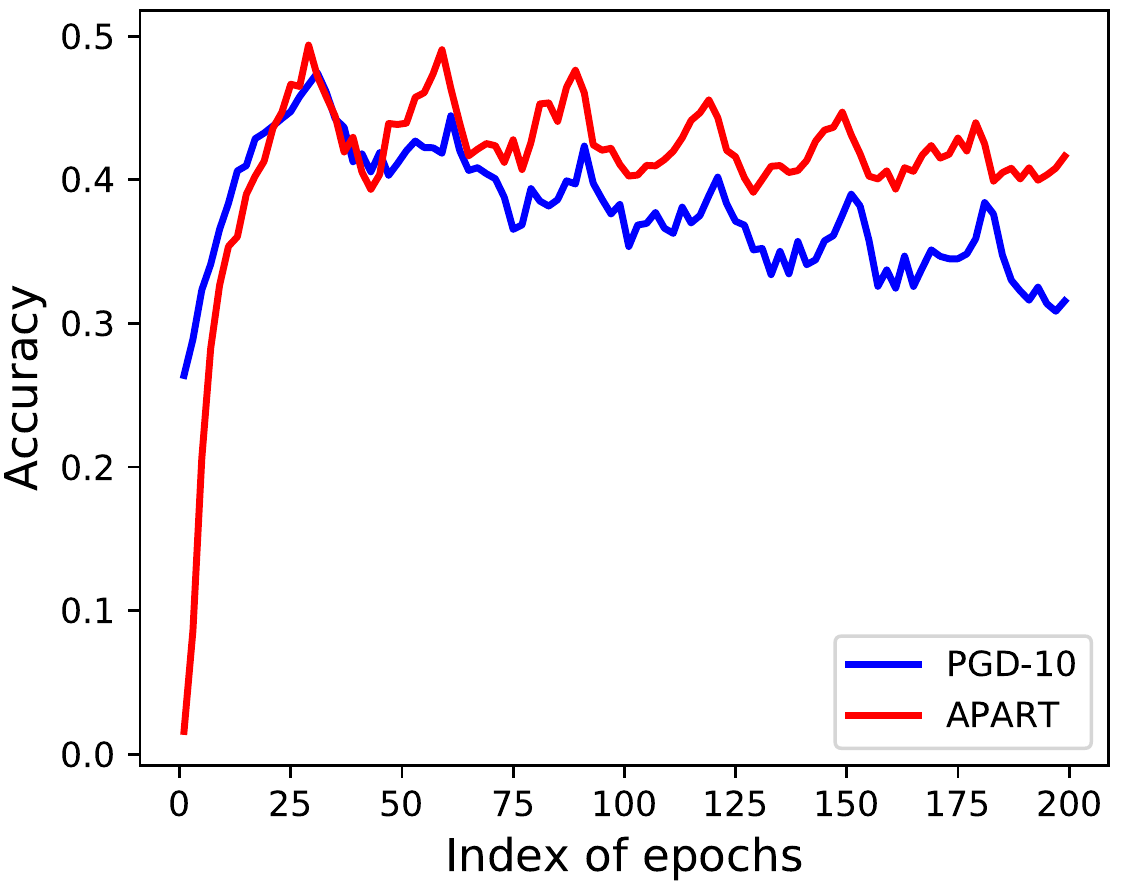}} \\

  \caption{
  Performance of PGD-10 and \our for Pre-ResNet18 on CIFAR-10. \our largely addresses the robustness drop, while its training is $\sim 4$ times faster than PGD-10 (as in Table~\ref{table:preact}).
}
	\label{fig:learning_curve}
\end{figure*}

Robustness drop has been widely observed after conducting adversarial training for too long.
For example, in Figure~\ref{fig:learning_curve}, the training objective of PGD-10 keeps decreasing, while the model robustness starts to drop after the 30th epoch. 
Remarkably, as in Figure~\ref{fig:fgsm_analysis}, stopping the FGSM training at the 21st epoch would yield a 0\% test robustness accuracy under the PGD-20 attack. 
Because of the performance gap between the robustness on the training set and the test set, this phenomenon has been long-regarded as overfitting.
However, its underlying mechanism remains mysterious. 
Here, we conduct systematic analyses to pursue its underpinnings. 

\subsection{Perturbation Strength Gap}

Comparing the robustness drops of FGSM and PGD, we observe that it leads to a smaller drop by employing a more complex perturbation generation process. 
Intuitively, since the robustness drop can be alleviated by using more powerful perturbations, the gap between these perturbation generation methods and ideal perturbations are too large. 
To further verify our intuition, we empirically estimate the perturbation strength.
For $\theta_{\mathcal{A}}^{(i)}$ and $f_{\cA}$, we approximate the perfect perturbation with a more powerful attack method $f_{\cB}$ and calculate the gap as $\cG_{\mathcal{A}\to\mathcal{B}}$ in Equation~\ref{eqn:cg_fgsm_pgd10}. 
\begin{equation}
\label{eqn:cg_fgsm_pgd10}
\cG_{\mathcal{A}\to\mathcal{B}}(\theta^{i}_\mathcal{A}) = \gL(\theta^{(i)}_{\mathcal{A}}; \xb+f_{\mathcal{B}}(\theta^{(i)}_{\mathcal{A}}, \xb, y), y) - \gL(\theta^{(i)}_{\mathcal{A}};\xb+\delta_{\mathcal{A}}, y).
\end{equation}

\begin{figure*}[t]
    \subfigure[$\cG(\cdot)$ by treating PGD-10 as perfect perturbations.]{
        \label{fig:pgd10_gap}
        \includegraphics[width=0.48\textwidth]{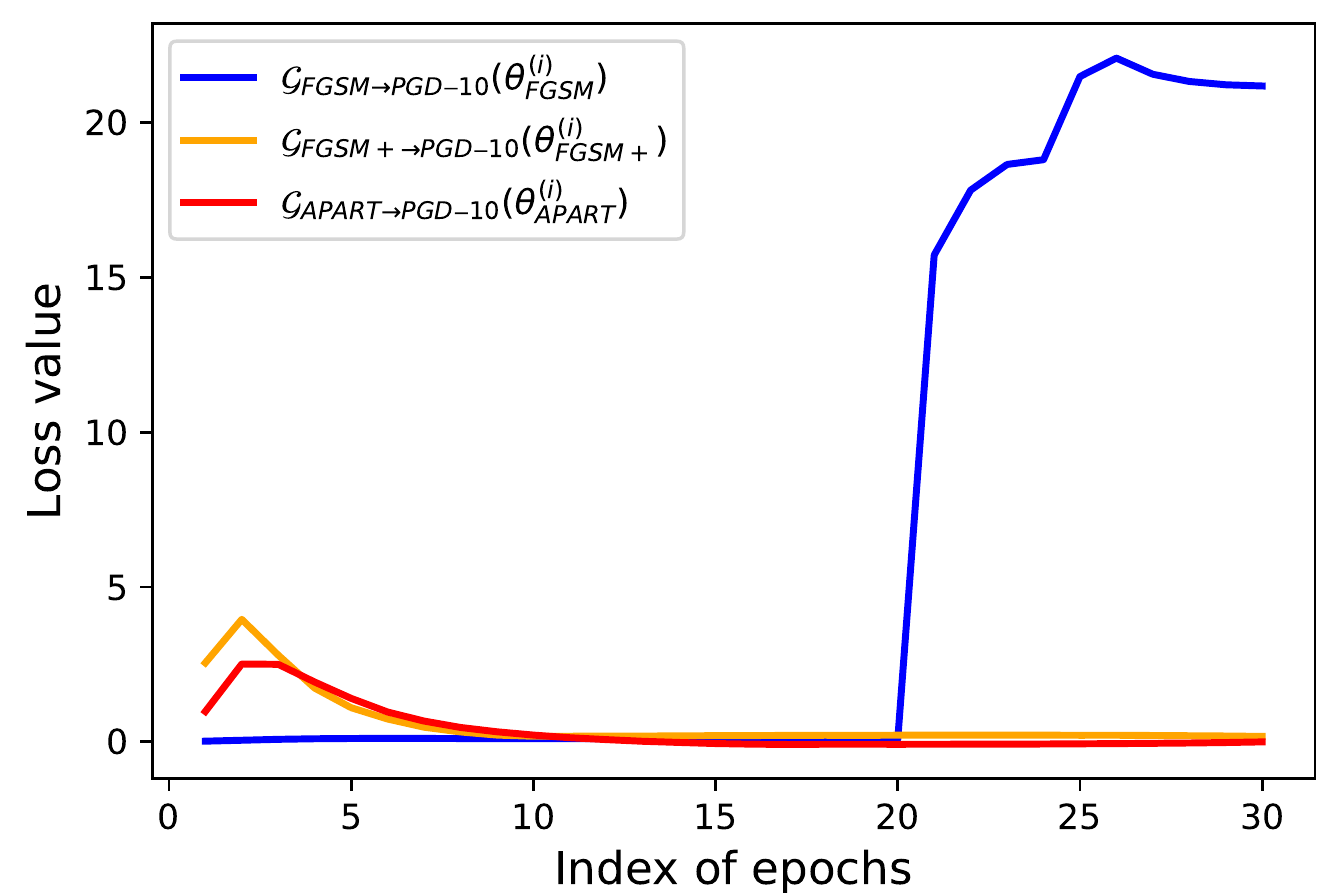}
    } 
	\subfigure[$\cG(\cdot)$ by treating PGD-20 as perfect perturbations.]{
      	\label{fig:pgd20_gap}
      	\includegraphics[width=0.48\textwidth]{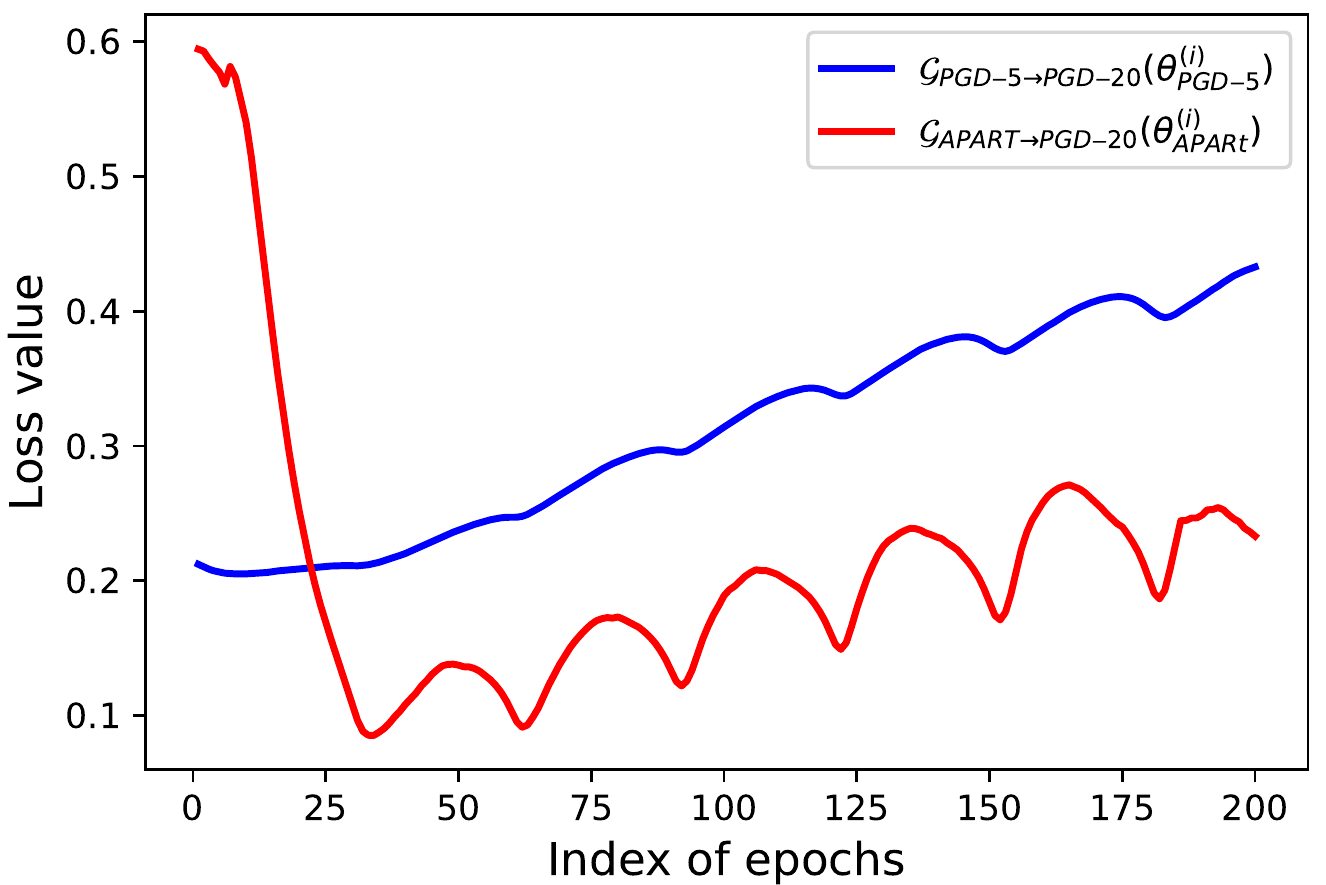}
	}
  \caption{
  Perturbation strength $\cG(\cdot)$ in different epochs. FGSM and PGD-5 suffer from the strength decrease while FGSM+ and \our are able to strengthen the generator progressively. 
  }
	\label{fig:diff_loss} 
\end{figure*}

The value of $\cG_{\fgsm\to\pgd{10}}$ is calculated with Pre-Act ResNet18 on the CIFAR-10 dataset, and the results are visualized in Figure~\ref{fig:diff_loss}(a). 
It shows that the strength gap is small in the early stage, dramatically explodes at the 21st epoch, and keeps a large value since then. 
Since this time point coincides with the time point for the robustness drop, it implies the strength gap to be the reason of the robustness drop. 
We also conduct empirical evaluations for PGD-5, visualize $\cG_{\pgd{5}\to\pgd{20}}$ in Figure~\ref{fig:diff_loss}(b), and observe a similar phenomenon---the gap keeps increasing during the training.

\subsection{Deterioration of FGSM}
\label{subsec:fgsm_deterioration}

To understand the significant robustness drop of FGSM, we further analyze the strength of FGSM-generated perturbations. 
First, we aim to explore whether FGSM-generated perturbations behave similar to random noise after its strength drop at the 21st epoch. 
Specifically, we test these perturbations, together with random noise, on a model that is separately trained with PGD-10.
We visualize the accuracy curves of $Acc(\theta_{\pgd{10}}, f_{\fgsm}(\theta^{(i)}_{\fgsm}, \cdot))$ and $Acc(\theta_{\pgd{10}}, f_{\random}(\cdot))$ in Figure~\ref{fig:defense}. 
The strength of perturbations increases in the early stage, and decreases dramatically later to the random noise level.
Also, this perturbation effectiveness drop happens at the 21st epoch, the same epoch where the significant robustness drop happens, which is another evidence verifying our intuition. 

Then, we aim to verify the intuition that allowing the perturbation generator to enhance itself can shield it from underfitting and alleviate the robustness drop. 
Intuitively, one straightforward way to strengthen generators is to parameterize them and update them together with model parameters. 
Specifically, we treat the perturbation initialization for the input (denoted as $\omega_\xb$) and the step size (referred as $\alpha$) as parameters of FGSM, and change the objective from Equation~\ref{eq:min} to Equation~\ref{eq:max-min}:
\begin{equation}
\label{eq:max-min}
    \max_{\alpha, \omega} \min_{\theta} \cL(\theta; \xb + f_{\fgsm}(\alpha, \omega_\xb; \theta, \xb, y), y).
\end{equation}
During model training, we update $\theta$ with gradient descent and update $\alpha$ and $\omega_\xb$ with gradient ascent.
We refer this variant as \textbf{FGSM+}.
We conduct experiments on Pre-Act ResNet18 on the CIFAR-10 dataset and visualize $Acc(\theta_{\fgsmp}, f_{\pgd{10}}(\theta^{(i)}_{\fgsmp}, \cdot))$ in Figure~\ref{fig:defense} as well as $\cG_{\fgsmp\to\pgd{10}}$ in Figure~\ref{fig:pgd10_gap}. 
Comparing to FGSM, FGSM+ largely alleviates the robustness drop, and the gap increase by parameterizing and updating the generator.
This not only supports our intuition that the perturbation underfitting contributes a lot to the robustness drop, but motivates us to shield generators by parameterizing and strengthening them with gradient updates.

\section{\our: Adaptive Adversarial Training}

Guided by our analyses, we propose adaptive adversarial training (APART), which further factorizes the perturbation for the input image into a series of perturbations, one for each residual block. 
Moreover, we employ different step sizes for different perturbations, treat them as learnable parameters, and update them to strengthen the perturbation generator. 

\subsection{Factorize the Input Perturbation as Multiple Layer-wise Perturbations}

For a multi-layer network, the perturbation generated at the input attacks not only the first layer, but also the following layers. 
Intuitively, existing methods like PGD-N implicitly blender these attacks with N additional forward- and backward-propagations on the input perturbation, which significantly inflates the computation cost. 
Here, we factorize the input perturbation as a series of perturbations and explicitly learn to combine them.
Specifically, we refer to the input of $i$-th residual block as $\xb_i$ and the output of $i$-th residual block as $\xb_i + \cnn(\xb_i)$. 
Then, we add the perturbation $\Delta \xb_i$ to the input of $\cnn(\cdot)$ to establish its robustness, \ie, $\xb_i + \cnn(\xb_i+\delta_i)$ where $\delta_i = \alpha_i \Delta \xb_i$. 
Although similar to PGD-N, this approach also involves multiple perturbations; they can be calculated with the same forward- and backward-propagations. 

To give the perturbation generator more flexibility to enhance itself, we also treat the perturbation initialization as learnable parameters. 
Since parameterizing the initialization consumes additional memory, we only parameterize the perturbation initialization at the input, and keep all other perturbations zero-initialized.
Thus,
the additional storage has roughly the same size with the dataset. 

\begin{algorithm}[ht]
\DontPrintSemicolon

\While{\emph{not converged}}{
    \For{$\xb, y$ \emph{in the training set}}{
        $\delta_1 \gets \alpha_{\omega} \cdot \omega_\xb$ \tcp{\footnotesize\hspace{-0.2cm}\fontdimen2\font=1ex initialize the perturbation for the model input.}
        
        \textcolor{red}{$\delta_i \gets \delta_i + \alpha_i \cdot \sign(\frac{\partial \cL(\theta; \xb + \delta_1, y)}{\partial \xb_i})$} \tcp{\footnotesize\hspace{-0.2cm}\fontdimen2\font=1ex calculate perturbations for block i.}
        
        $\delta_1 \gets \max(\min(\delta_{1}, -\epsilon), +\epsilon)$ \tcp{\footnotesize\hspace{-0.2cm}\fontdimen2\font=1ex project $\delta_1$ to the feasible set.}
        
        \textcolor{blue}{$\theta = \theta - \mu_\theta \cdot \frac{\partial \cL(\theta; \{\xb_i\}_{i=1}^n + \{\delta_i\}_{i=1}^n, y)}{\partial \theta}$} \tcp{\footnotesize\hspace{-0.2cm}\fontdimen2\font=1ex update parameters.}
        
        \textcolor{blue}{$\omega_\xb = \max(\min(\omega_\xb + \mu_\omega \cdot \sign(\frac{\partial \cL(\theta; \{\xb_i\}_{i=1}^n + \{\delta_i\}_{i=1}^n, y)}{\partial \omega_\xb}), -1), 1)$} \tcp{\footnotesize\hspace{-0.2cm}\fontdimen2\font=1ex update $\omega_\xb$ and project it to the scale of the gradient sign.}
        
        \textcolor{blue}{$\alpha_i = \alpha_i + \mu_\alpha \cdot (\frac{\partial \cL(\theta; \{\xb_i\}_{i=1}^n + \{\delta_i\}_{i=1}^n, y)}{\partial \alpha_i} - \lambda \cdot \frac{|\alpha_i|_2^2}{\partial \alpha_i})$} \tcp{\footnotesize\hspace{-0.2cm}\fontdimen2\font=1ex update step sizes.}
        
    }
}
\Return{$\theta $}
\caption{\our (the \textcolor{red}{first} and the \textcolor{blue}{second} round forward- and backward-propagations are marked with \textcolor{red}{red} and \textcolor{blue}{blue}, respectively; $\epsilon$ is the perturbation bound; $\mu_\theta$, $\mu_\omega$, and $\mu_\alpha$ are learning rates for $\theta$, $\omega$, and $\alpha$ respectively; other notations are summarized in Table~\ref{table:notation}).}
\label{algo:adapt}
\end{algorithm}

\subsection{\our Algorithm}

We summarize \our in Algorithm~\ref{algo:adapt}, which contains two rounds of forward- and backward-propagations. 
In the first round, it initializes the input perturbation and calculates gradients for
both the input of the first layer and the input of all the following blocks. 
In the second round, it applies the generated perturbations to the input of the corresponding blocks, \ie, change $\xb_i + \cnn(\xb_i)$ to $\xb_i + \cnn(\xb_i+ \delta_i)$. 
Then, besides updating model parameters with gradient descent, we enhance the generator with gradient ascent (\ie, updating step sizes $\alpha_i$ and the perturbation initialization $\omega_\xb$). 
Note that, to control the magnitude of step sizes $\alpha_i$, we add a $L_2$ regularization to its updates and use $\lambda$ to control it (as line 8 in Algorithm~\ref{algo:adapt}). 

Note that, calculating the exact gradients of $\omega_\xb$ or $\alpha_\omega$ requires a second order derivation ($\Delta \omega_\xb$ and $\Delta \alpha_\omega$ are based on $\delta_i$, and the calculation of $\delta_i$ includes some first order derivations involving $\omega_\xb$ and $\alpha_\omega$). 
Due to the success of the First-order MAML (FOMAML)~\citep{Finn2017ModelAgnosticMF}, we simplifies the calculation by omitting higher order derivations. 
Specifically, FOMAML demonstrates the effectiveness to ignore higher order derivations and approximate the exact gradient with only first order derivations. 
Here, we have a similar objective with FOMAML---FOMAML aims to find a good model initialization, and we try to find a good perturbation initialization.
Thus, we also restrict gradient calculations to first-order derivations.
In this way, \our has roughly the same complexity with FGSM and it is significantly faster than PGD-N. 

If only adding perturbations to the input image, \our would become FGSM+ (as introduced in Section~\ref{subsec:fgsm_deterioration}). 
Here, we further factorize the input perturbation and give the generator more flexibility to be strengthened during the training. 
As in Figures~\ref{fig:diff_loss}~and~\ref{fig:learning_curve}, \our largely shields adversarial training from robustness drop.
Systematic evaluations are further conducted as below.

\section{Experiments}
\label{sec:exp}

\subsection{Experimental Settings}

\begin{table}[t]
  \caption{Model Performance of Pre-ResNet18 on the CIFAR-10 dataset.}
  \label{table:preact}
\tablespace
  \centering
  \begin{tabular}{lccccc}
    \toprule
     Methods & Clean Data & PGD-20 Attack & C\&W Attack & Gaussian Noise & Time/Epoch\\
    \midrule
    ATTA-10 & 82.10\% & 49.03\% & 58.30\% & 73.10\% & 140 secs\\
    PGD-10 & 82.05\% &  47.43\% & 56.55\% & 72.33\% & 133 secs\\
    Free-8 & 81.64\% & 47.37\% & 56.10\% & 72.56\% &62 secs\\
    F+FGSM & {83.81\%} & 46.06\% & 55.34\% & 72.25\% & {20} secs\\
    \our & 82.45\% & {49.30\%} & {58.50\%} & {73.18\%} & 29 secs\\
    \bottomrule
  \end{tabular}
\end{table}

\begin{table}[t]
  \caption{Model Performance of WideResNet34-10 on the CIFAR-10 dataset.}
  \label{table:cifar10, wide}
\tablespace
  \centering
  \begin{tabular}{lccccc}
    \toprule
     Methods & Clean Data &  PGD-20 Attack  & C\&W Attack & Gaussian Noise & Time/Epoch \\
    \midrule
    ATTA-10 & 83.80\% & 54.33\% &  59.11\% & 77.05\% & 706 secs \\
    PGD-10 & 86.43\% & 47.12\% & 56.45\% & 75.70\% &680 secs \\
    Free-8 & 85.54\% & 47.68\% & 56.31\% & 75.98\% &252 secs \\
    F+FGSM & 85.10\% & 46.37\% &  56.21\% & 75.10\% & 122 secs \\
    \our & 84.75\%  & 51.10\% &  58.50\% & 76.50\% & 162 secs\\

    \bottomrule
  \end{tabular}
\end{table}

\smallsection{Datasets}
We conduct experiments on the CIFAR-10 and CIFAR-100 datasets~\citep{Krizhevsky2009LearningML} as well as the ImageNet dataset~\citep{krizhevsky2012imagenet}.

\smallsection{Neural Architectures}
To show the generality of our \our framework, we conduct experiments with Pre-ResNet~\citep{he2016identity} and WideResNet~\citep{Zagoruyko2016WideRN}.
Specifically, we use Pre-ResNet18 and WideResNet34-10 on the CIFAR-10 dataset, Pre-ResNet18 on the CIFAR-100 dataset, and ResNet50~\citep{he2016deep} on the ImageNet dataset. 

\smallsection{Data Augmentation}
For the CIFAR datasets, we apply random flipping as a data augmentation procedure and take a random crop with $32 \times 32$ from images padded by $4$ pixels on each side~\citep{Lee2015DeeplySupervisedN}. 
For the ImageNet dataset, we divide the training into three phases, where phases 1 and 2 use images resized to 160 and 352 pixels and the third phase uses the whole images~\citep{Wong2020FastIB}.

\smallsection{Optimizer}
For all experiments on the CIFAR datasets, we use the SGD solver~\citep{Robbins2007ASA} with momentum of 0.9 and train for 30 epochs with cyclic learning rate decay for $\mu_\theta$, where the maximum learning rate is 0.2 and minimum learning rate is 0.
Cyclic learning rate scheduler~\citep{Smith2017CyclicalLR} is adopted to help convergence of neural networks and reduce the number of learning rate trials. 
Specifically, the learning rate starts from zero, reaches to the maximum learning rates and decreases to zero. 
For the ImageNet dataset, we adopt a setting similar to~\cite{Wong2020FastIB} and train the model for 15 epochs; The maximum learning rate of cyclic learning rate schedule is 0.4 and the minimum one is 0. 

\smallsection{Other Hyper-parameters}
For all experiments, we apply the cyclic learning rate scheduler for $\mu_\alpha$. 
On the CIFAR datasets, the maximum learning rate is $5\times10^{-8}$ and $\lambda$ is set as $400$. On the ImageNet dataset, we set the maximum learning rate as $4\times10^{-9}$ and $\lambda$ as $5000$.
Due to the similarity between line 4 and line 7 in Algorithm~\ref{algo:adapt}, we set $\mu_{\omega}$ as $\frac{\alpha_1}{\alpha_{\omega}}$, which makes the update on $\omega$ has a similar impact with the update in line 4 of Algorithm~\ref{algo:adapt}. 

\smallsection{Attacks for Robustness Evaluation}
We adopt PGD-20~\citep{Madry2017TowardsDL}, Gaussian random noise, and C\&W~\citep{Carlini2017TowardsET} as the attack methods for evaluation. 
For both adversarial training and evaluation, we restrict perturbations to  $|\delta|_{\infty}\leq8/255$ on the CIFAR datasets, $|\delta|_{\infty}\leq2/255$ on the ImageNet dataset.

\smallsection{Infrastructure}
Our experiments are conducted with NVIDIA Quadro RTX 8000 GPUs, and mixed-precision arithmetic~\citep{Micikevicius2018MixedPT} is adopted to accelerate model training. 
The training speed of our method or baselines is evaluated on an idle GPU. 

\subsection{Compared Methods}
We select four state-of-the-art adversarial training methods for comparison. 
To ensure a fair comparison on efficiency, we report accuracy and training time based on our experiments on the CIFAR datasets. 
As to ImageNet, we directly refer to the number reported in the original papers.
\begin{itemize}[nosep, leftmargin=*]
    \item \textbf{PGD-N}\citep{Madry2017TowardsDL} is a classical, sophisticated adversarial training method. PGD is an iterative version of FGSM with uniform random noise as initialization and N is the number of iterations.
    \item \textbf{Free-m}~\citep{shafahi2019adversarial} updates both the model parameters and image perturbations with one simultaneous backward pass instead of separate computations by training on the same minibatch $m$ times in a row. Here we set $m = 8$. 
    \item \textbf{ATTA-K}~\citep{Zheng2019EfficientAT} uses the adversarial examples from neighboring epochs. K is the number of iterations and denotes the strength of attack. 
    \item \textbf{F+FGSM}~\citep{Wong2020FastIB} uses a large step size and random initialization to improve the performance of FGSM to be comparable to that of PGD-10.
\end{itemize}

\subsection{Performance Comparison}

\smallsection{CIFAR-10}
We summarize the model performance in Table~\ref{table:preact} and Table~\ref{table:cifar10, wide}. 
For Pre-ResNet18, \our improves adversarial accuracy over 
PGD-10 with 4.6x speedup. 
For example, as PGD-10 leads to a $47.43\%$ accuracy under the PGD-20 attack, \our achieves a $49.26\%$ accuracy. 
Among all methods, \our achieves the highest adversarial accuracy under PGD-20, Gaussian random noise, and C\&W attack, with a satisfying performance on clean images. 
Similar pattern is observed for WideResNet34-10, \our outperforms PGD-10 under various attacks with a 4.2x speedup, at the cost of a bit clean accuracy. 

\smallsection{CIFAR-100}
We summarize model performance in Table~\ref{table:cifar100}. 
Similar to CIFAR-10, \our is not only faster than PGD-10 (4.6x speedup), but also has more robustness (more than 1\% absolute accuracy). 
It is worth mentioning that \our achieves the highest performance on corrupted images, the 2rd highest performance on clean images, and the 2rd fastest training speed. 

\smallsection{ImageNet}
We summarize model performance in Table~\ref{table:ImageNet}.
Among all the compared methods, ATTA-2 reaches the highest accuracy under PGD-10 attack with the highest computation cost. 
\our achieves a similar performance as ATTA-2, but with a much less training cost. 


\begin{table}[t]
  \caption{Model Performance of Pre-ResNet18 on the CIFAR-100 dataset.}
  \label{table:cifar100}
\tablespace
  \centering
  \begin{tabular}{lccccc}
    \toprule
     Methods & Clean Data & PGD-20 Attack & C\&W Attack & Gaussian Noise & Time/Epoch\\
    \midrule
    ATTA-10 & 56.20\% & 25.60\% & 30.75\% & 42.10\% & 140 secs\\
    PGD-10 & 55.14\% &  25.61\%  &30.65\%  &  42.41\% & 133 secs\\
    Free-8 & 55.13\% & 25.88\% & 30.55\% &42.15\% &62 secs\\
    F+FGSM & 57.95\% & 25.31\% & 30.06\% & 42.03\%& {20} secs \\
    \our & 57.10\% & {27.10}\% & 32.36\% & 43.30\%& 29 secs\\
    
    \bottomrule
  \end{tabular}
\end{table}

\begin{table}[t]
  \caption{Model Performance of ResNet50 on the ImageNet dataset. Total training time is reported. $^*$ indicates the time is not directly comparable due to hardware differences. $^{\tiny+}$ indicates FP32 training.}
\tablespace
  \label{table:ImageNet}
  \centering
  \begin{tabular}{lccc}
    \toprule
     Methods & Clean Data & PGD-10 Attack & Time(total)\\
    \midrule
    ATTA-2~\citep{Zheng2019EfficientAT} & 60.70\% & 44.57\% & 97$^{*{\tiny+}}$ hrs\\
    Free-4~\citep{shafahi2019adversarial} & 64.44\% & 43.52\%  & 39$^*$ hrs\\
    F+FGSM~\citep{Wong2020FastIB} & 60.90\% & 43.46\% & 12$^*$ hrs\\
    \our & 60.52\% & 44.30\% & 15$^{\;\,}$ hrs\\
    \bottomrule
  \end{tabular}
\end{table}

\subsection{Balance between Clean Accuracy and Accuracy}
As shown in Tables~\ref{table:preact}, \ref{table:cifar10, wide}, and \ref{table:cifar100}, the performance on corrupted images have consistent trends across different models, \eg, if method $\cA$ outperforms $\cB$ under PGD-20 attack, $\cA$ likely also outperforms $\cB$ under C\&W attack or Gaussian noise. 
At the same time, it seems that the robustness improvement usually comes at the cost of accuracy on clean images. 
To better understand the trade-off between the clean accuracy and the model robustness, we employ different $\epsilon$ values during training (i.e., $[\frac{2}{255}, \frac{3}{255}, \cdots, \frac{10}{255}]$), train multiple models for each method, and visualize their performance of Pre-ResNet20 on the CIFAR-10 dataset in Figure~\ref{fig:diff_eps}.
Points of \our locate in the right top corner of the figure and significantly outperform other methods. 
This further verifies the effectiveness of \our.

\subsection{Ablation Studies}
\label{subsec:exp_ablation}
\our employs two techniques to parameterize the perturbation generator.
The first is to learn an initialization for the perturbation, and the second is to factorize input perturbations into a series of perturbations. 
To understand the effectiveness of them, we conduct an ablation study and summarized the results in Table~\ref{table:part}.
Removing layer-wise perturbations leads to a 2.95\% drop on accuracy under PGD-20 attack, and removing perturbation initialization leads to a 2.00\% drop.
Therefore, both techniques are helpful and necessary to achieve a better model robustness.
\begin{table}[t]
  \caption{Ablation study of \our on the CIFAR-10 dataset with Pre-ResNet20.}
  \vspace{2mm}
  \label{table:part}
  \centering
  \begin{tabular}{lcc}
    \toprule
    Training Methods & Clean Data & PGD-20 Attack \\
    \midrule
    \our & 82.45\% & 49.30\% \\ 
    \our(w/o layer-wise perturbation) & 83.30\% & 46.35\%\\
    \our(w/o perturbation initialization) & 82.00\% & 48.40\%\\
    \bottomrule
  \end{tabular}
\end{table}

\subsection{Evolution of Step Sizes in Perturbation Composing}
\begin{figure}[t]
\begin{minipage}[t]{0.55\textwidth}
\centering
\includegraphics[height=1.55in]{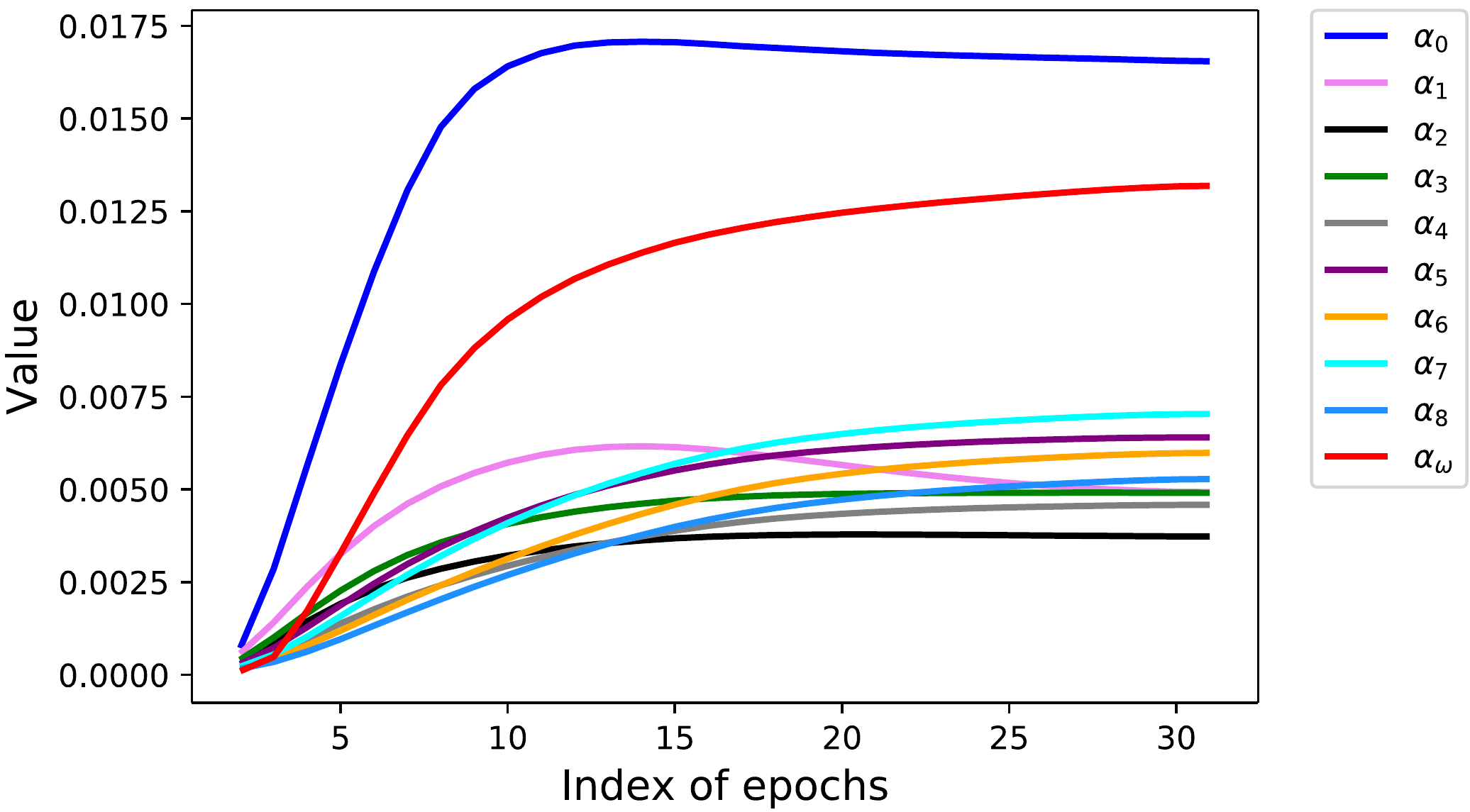}
\caption{Step sizes in different epochs \newline 
(with Pre-ResNet18 on CIFAR-10)}
\label{fig:eps}
\end{minipage}%
\begin{minipage}[t]{0.4\textwidth}
\includegraphics[height=1.55in]{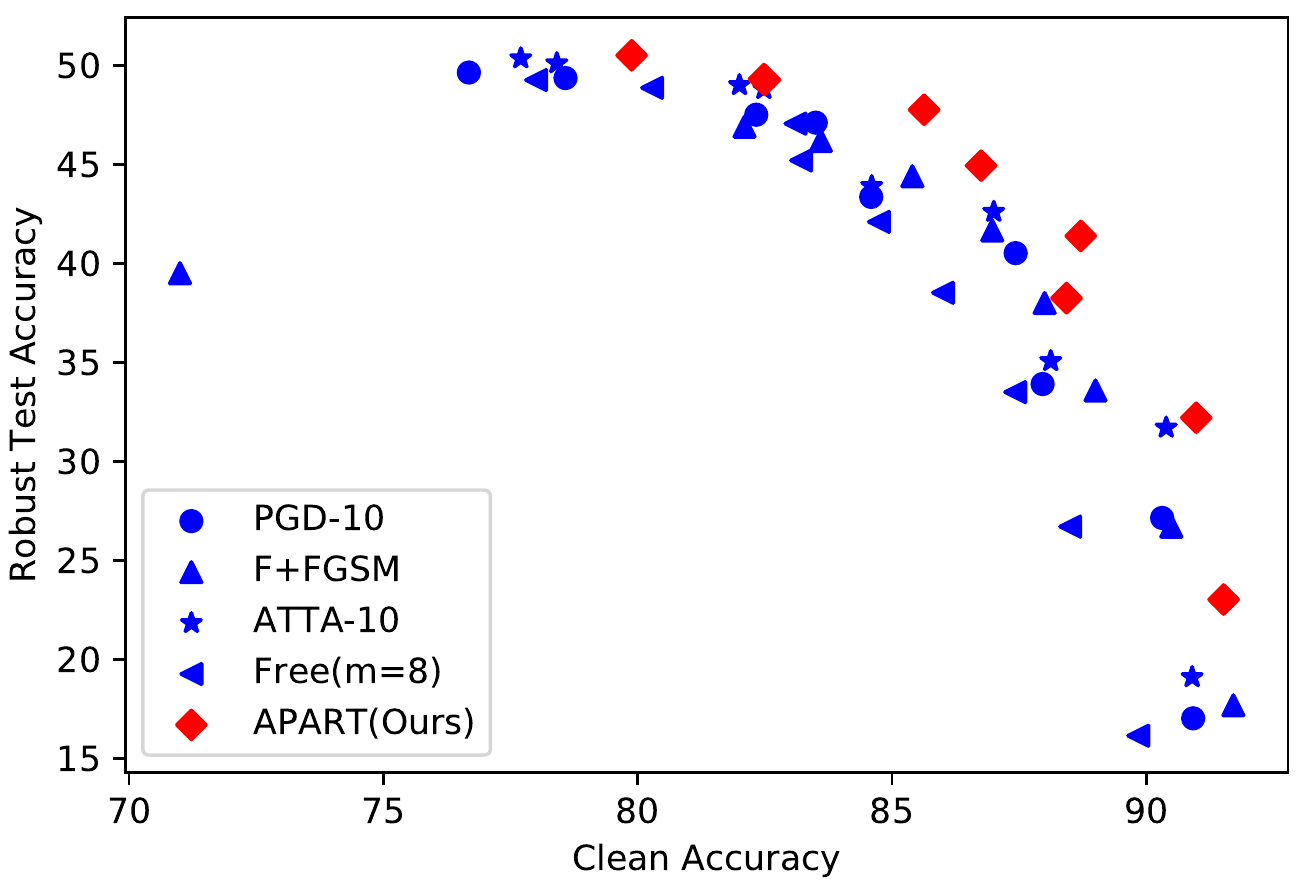}
\caption{Pre-ResNet18 performance on CIFAR-10 ($\epsilon_{train}=[\frac{2}{255}, \cdots, \frac{10}{255}]$)}
\label{fig:diff_eps}
\end{minipage}
\end{figure}

After factorizing the input perturbation as a series of perturbations, we employ learnable step sizes to compose perturbations more effectively. 
To better understand these step sizes, we visualize their values during the training of Pre-ResNet20 on the CIFAR-10 dataset in Figure~\ref{fig:eps}.
It shows that the first-layer perturbation is more important than others.
Also, the step sizes of perturbation at the first and the second layers decrease after 10 epochs, while other step sizes keep increasing across the training.
This phenomenon verifies that \our is able to adapt the generator setting to different training stages. 

\section{Related Work}
\subsection{Adversarial Training}
Adversarial training is a common method to defend adversarial attacks. \cite{goodfellow2014explaining} recognize the cause of the adversarial vulnerability to be the extreme nonlinearity of deep neural networks and introduced the fast gradient sign method (FGSM) to generate adversarial examples with a single gradient step. \cite{Madry2017TowardsDL} propose an iterative method based on FGSM with random starts, Projected Gradient Descent (PGD). PGD adversarial training is effective but time-consuming, and thus some recent work also pays attention to the efficiency of adversarial training. For example, \cite{shafahi2019adversarial} propose to update both the model parameters and image perturbations using one simultaneous backward pass. \cite{Zhang2019YouOP} show that the first layer of the neural network is more important than other layers and make the adversary computation focus more on the first layer. \cite{Zheng2019EfficientAT} also improve the utilization of gradients to reuse perturbations across epochs. \cite{Wong2020FastIB} use uniform random initialization to improve the performance of FGSM adversarial training. 
\our improves the efficiency and effectiveness of adversarial training by factorizing the input perturbation as a series of perturbations. Previous methods only added the perturbation to input images, while \our adds perturbation to the input of residual blocks. Perturbations added to intermediate variables help improve the robustness, as discussed in Section~\ref{subsec:exp_ablation}.

\subsection{Robustness Drop}
\cite{Wong2020FastIB} mention the robustness drop as overfitting and first identify a failure mode named as ``catastrophic overfitting'', which caused FGSM adversarial training to fail against PGD attacks. \cite{Rice2020OverfittingIA} further explore the overfitting in other adversarial training methods, such as PGD adversarial training and TRADES. They observe that the best test set performance was achieved after a certain epochs and further training would lead to a consistent decrease in the robust test accuracy, and therefore explain it as ``robust overfitting''. \cite{Rice2020OverfittingIA} show that robustness drop is a general phenomenon but they did not analyze its cause. 
In this work, we explore the nature of robustness drop in adversarial training and further propose \our to address the perturbation underfitting issue.

\section{Conclusion}
\label{sec:conclusion}

In this paper, we explore to answer the question of \emph{why the robustness drops after conducting adversarial training for too long}.
As the common wisdom views this phenomenon as overfitting, our analyses in Section~\ref{sec:overfit} suggest that the primary cause of the robustness drop is perturbation underfitting. 
In the light of our analyses, we propose \our, an \underline{a}da\underline{p}tive \underline{a}dve\underline{r}sarial \underline{t}raining framework, which parameterizes perturbation generation and progressively strengthens them.
Specifically, \our parameterizes the perturbation initialization and factorizes the input perturbation into a series of perturbations, one for each layer in the neural networks.
Shielding perturbations from underfitting unleashes the potential of our framework to achieve a better performance. 
\our achieves comparable or even better performance than PGD-10 with less than 1/4 of its training cost. 

Our work opens up new possibilities to better understand adversarial training and adversarial vulnerability. 
It leads to many interesting future works, including discussing the balance between accuracy on corrupted images and clean accuracy, applying our proposed \our to other architectures like Transformer~\citep{liu2020admin}, and explicitly regularize the perturbation generator strength gap. 

\section*{Acknowledge}

The research was sponsored in part by DARPA No. W911NF-17-C-0099 and No.  FA8750-19-2-1004, National Science Foundation CA-2040727, IIS-19-56151, IIS-17-41317, IIS 17-04532, and IIS 16-18481, and DTRA HDTRA11810026. Any opinions, findings, and conclusions or recommendations expressed herein are those of the authors and should not be interpreted as necessarily representing the views, either expressed or implied, of the U.S. Government. The U.S. Government is authorized to reproduce and distribute reprints for government purposes notwithstanding any copyright annotation hereon..

\bibliography{iclr2021_conference}
\bibliographystyle{iclr2021_conference}

\end{document}